\documentclass[runningheads]{llncs}

\usepackage{eccv}

\usepackage{eccvabbrv}

\usepackage{graphicx}
\usepackage{booktabs}

\usepackage[accsupp]{axessibility}  

\usepackage[pagebackref,breaklinks,colorlinks,citecolor=eccvblue]{hyperref}

\usepackage{orcidlink}

\begin{document}

\title{GeoTransfer: Generalizable Few-Shot Multi-View Reconstruction via Transfer Learning} 

\titlerunning{GeoTransfer}

\author{Shubhendu Jena \and
Franck Multon \and
Adnane Boukhayma}

\authorrunning{S.~Jena et al.}

\institute{Inria, Univ. Rennes, CNRS, IRISA, M2S, France}

\maketitle

\vspace{-15pt}
\begin{abstract}
This paper presents a novel approach for sparse 3D reconstruction by leveraging the expressive power of Neural Radiance Fields (NeRFs) \cite{mildenhall2021nerf} and fast transfer of their features to learn accurate occupancy fields. Existing 3D reconstruction methods from sparse inputs still struggle with capturing intricate geometric details and can suffer from limitations in handling occluded regions. On the other hand, NeRFs \cite{mildenhall2021nerf} excel in modeling complex scenes but do not offer means to extract meaningful geometry. Our proposed method offers the best of both worlds by transferring the information encoded in NeRF \cite{mildenhall2021nerf} features to derive an accurate occupancy field representation. We utilize a pre-trained, generalizable state-of-the-art NeRF network \cite{johari2022geonerf} to capture detailed scene radiance information, and rapidly transfer this knowledge to train a generalizable implicit occupancy network. This process helps in leveraging the knowledge of the scene geometry encoded in the generalizable NeRF \cite{johari2022geonerf} prior and refining it to learn occupancy fields, facilitating a more precise generalizable representation of 3D space. The transfer learning approach leads to a dramatic reduction in training time, by orders of magnitude (\ie from several days to 3.5 hrs), obviating the need to train generalizable sparse surface reconstruction methods from scratch. Additionally, we introduce a novel loss on volumetric rendering weights that helps in the learning of accurate occupancy fields, along with a normal loss that helps in global smoothing of the occupancy fields. We evaluate our approach on the DTU dataset \cite{aanaes2016large} and demonstrate \textbf{state-of-the-art performance} in terms of reconstruction accuracy, especially in challenging scenarios with sparse input data and occluded regions. We furthermore demonstrate the generalization capabilities of our method by showing qualitative results on the Blended MVS \cite{yao2020blendedmvs} dataset without any retraining. Project page : \href{https://shubhendu-jena.github.io/geotransfer/}{https://shubhendu-jena.github.io/geotransfer/}
  \vspace{-5pt}
\end{abstract}

\section{Introduction}
\label{sec:intro}

Creating three-dimensional structures from a set of images is a fundamental task in the realm of computer vision, finding broad applications in fields like robotics, augmented reality, and virtual reality. The first seminal deep learning approaches tackling this problem used Multi-view stereo techniques (MVS), as demonstrated by MVSNet \cite{yao2018mvsnet} and its successors \cite{yang2020cost, gu2020cascade, wang2021patchmatchnet, ding2022transmvsnet}. These methods build 3D cost volumes based on the camera frustum, deviating from traditional euclidean space, to achieve accurate depth-map estimation. However, they often necessitate subsequent steps, such as depth-map filtering, fusion, and mesh reconstruction, and exhibit susceptibility to noise, texture-less regions, and gaps. \\
Unlike seminal work relying on explicit representations (\eg meshes \cite{wang2018pixel2mesh,kato2018neural,jena2022neural} and point clouds \cite{fan2017point,aliev2020neural,kerbl20233d}), neural implicit reconstruction methods \cite{yariv2021volume, wang2021neus, oechsle2021unisurf, niemeyer2020differentiable, darmon2022improving} constitute another popular class of strategies to address this challenge, creating precise and realistic geometry from multi-view images through the use of volume rendering and neural implicit representations based on the Sign Distance Function (SDF) \cite{park2019deepsdf} and its variations. However, despite their effectiveness, these approaches come with inherent limitations, including a lack of cross-scene generalization capabilities and the need for substantial computational resources to train them from scratch for each scene. Moreover, these techniques heavily depend on a large number of input views. However, due to many constrained scenarios (\eg out-of-the-studio, low budget, \etc) and in the interest of wider applicability, there is active interest in seeking solutions that can deliver under minimal input. \\
To solve these issues, recent investigations, in the context of novel-view synthesis \cite{johari2022geonerf,yang2023freenerf,niemeyer2022regnerf} and 3D reconstruction \cite{ren2023volrecon,long2022sparseneus,liang2024retr} have sought solutions by relying on learned data priors across many training scenes, by conditioning the implicit representation on spatially local features obtained from the sparse input images through generalizable encoders. This approach has proven effective in achieving remarkable cross-scene generalization capabilities, even with sparse views as input. Most relevant to our approach is GeoNeRF \cite{johari2022geonerf} which constructs a cost volume to enable geometry-aware scene reasoning, followed by attention-based view aggregation and volumetric rendering to learn the radiance field of a scene.\\
In this paper, differently from concurrent works, we explore the idea of using this pre-existing generalizable state-of-the-art NeRF to obtain scene reconstructions through transfer learning. In this context, we show that under the assumption of our scene being composed of solid, non-transparent objects, it is possible to rapidly transform the generalizable sampling-dependent opacity obtained from the density field of GeoNeRF \cite{johari2022geonerf} to a generalizable sampling-independent occupancy. This strategy also leads to a drastic reduction of training time from the order of several days to a couple of hours, which removes the need to train generalizable sparse reconstruction methods \cite{ren2023volrecon,long2022sparseneus,liang2024retr} from scratch. We introduce a novel volumetric rendering weight loss, in addition to a surface normal based smoothing loss to further refine our occupancy field, leading to the current state-of-the-art results in sparse 3D reconstruction on the DTU dataset \cite{aanaes2016large} without requiring any test-time optimization, and outperforming in this process the current state-of-the-art generalizable SDF based 3D reconstruction networks \cite{ren2023volrecon,long2022sparseneus,liang2024retr}.
\\
In summary, our contribution can be summarized as:
\begin{itemize}
  \item We explore a novel strategy of fast adaptation of an existing state-of-the-art generalizable NeRF method to obtain a generalizable occupancy network by transferring and fine-tuning its features. This yields the \textbf{state-of-the-art performance on DTU \cite{aanaes2016large}} reconstruction from sparse views \ref{tab:table1}.
  \item The decrease in training duration from multiple days to just a few hours, all the while achieving state-of-the-art performance, removes the necessity to train computationally intensive sparse generalizable surface reconstruction techniques from scratch.
  \item Among the losses we use for our transfer learning framework, we propose a novel volumetric rendering weight loss to impose the properties followed by an ideal occupancy field in a volumetric rendering framework which leads to the learning of a more accurate occupancy functions (\ref{fig:weight_dist}). 
\end{itemize}

\section{Related Work}

There is a substantial body of work on the subject of 3D reconstruction, and we review here work we deemed most relevant to the context of our contribution.\\ 
\\{\bf Neural Surface Reconstruction.} In the realm of neural surface reconstruction, the utilization of neural implicit representations enables the depiction of 3D geometries as continuous functions that can be computed at arbitrary spatial locations. Due to their capability to represent complex and detailed shapes in a compact and efficient manner, these representations demonstrate significant potential in tasks such as 3D reconstruction \cite{younes2024sparsecraft, darmon2022improving, jiang2020sdfdiff, kellnhofer2021neural, niemeyer2020differentiable, oechsle2021unisurf, wang2021neus, yariv2021volume, yariv2020multiview, yu2022monosdf}, shape representation \cite{atzmon2020sal, gropp2020implicit, mescheder2019occupancy, park2019deepsdf}, and novel view synthesis \cite{liu2020neural, mildenhall2021nerf, trevithick2021grf}. The emergence of NeRF \cite{mildenhall2021nerf} has instigated a significant shift in the paradigm towards employing similar techniques for these tasks. IDR \cite{yariv2020multiview} utilizes surface rendering to acquire geometry from multi-view images but necessitates additional object masks. Unisurf \cite{oechsle2021unisurf}, which is most relevant to our work, models the local opacity of a NeRF \cite{mildenhall2021nerf} with an occupancy network, which allows them to train on multiple datasets including ones involving forward-facing scenes such as LLFF \cite{mildenhall2019local}. Differently, several methods have attempted to rewrite the density function in NeRF \cite{mildenhall2021nerf} using Signed Distance Function (SDF) and its variants, successfully achieving plausible geometry. Significantly, NeuS \cite{wang2021neus} formulates an unbiased and occlusion aware volumetric weight function equation by employing logistic sigmoid functions. Conversely, Volsdf \cite{yariv2021volume} incorporates a signed distance function into the density formulation and introduces a sampling strategy that meets a determined error bound on the transparency function. HF-NeuS \cite{wang2022hf} improves upon NeuS \cite{wang2021neus} by modeling transparency as a transformation of the estimated signed distance field and proposes to decompose the signed distance function into a base function and a displacement function with a coarse-to-fine strategy to gradually increase the high-frequency details. These methods offer a robust approach for multi-view 3D reconstruction from 2D images. However, these methods require prolonged optimization for training each scene independently and also require a substantial number of dense images, making it challenging to generalize to unknown scenes and limiting deployment. 
\\{\bf Generalizable NeRFs.} Some recent methods \cite{deng2022depth, niemeyer2022regnerf, wynn2023diffusionerf, jain2021putting, kim2022infonerf, li2023regularizing, younes2024sparsecraft} synthesize novel views on a single scene with sparse views, albeit facing difficulties in understanding the underlying geometry of the scene, which they attempt to solve using several geometry-based regularization strategies \cite{Ouasfi_2024_CVPR,ouasfi2024fewshot,gropp2020implicit,ben2022digs}. To attempt to solve this problem, certain methods \cite{chen2021mvsnerf, johari2022geonerf, chibane2021stereo, liu2022neural, wang2021ibrnet, li2023learning, yu2021pixelnerf} generate novel views in unknown scenarios through a generalization approach, constructing neural radiance fields on sparse views. These methods can infer on unknown scenarios without fine-tuning after training on multiple known scenarios, which involves incorporating priors derived from a larger model trained on multi-view image datasets. Among such methods, PixelNeRF \cite{yu2021pixelnerf} conditions itself on features extracted by a CNN. MVSNeRF \cite{chen2021mvsnerf} constructs a neural volume from the cost volume obtained by warping image features and conditions itself on this neural volume. IBRNet \cite{wang2021ibrnet} aggregates features from nearby views to infer geometry and adopts an image-based rendering approach. NeuRay \cite{liu2022neural} utilizes neural networks to model and handle occlusions, thereby improving the quality and accuracy of image-based rendering. GeoNeRF \cite{johari2022geonerf}, another recent method and one that we build on, employs a cascaded cost volume and an attention-based technique to aggregate information from different views. However, deriving the scene geometry from the volume density of these NeRF \cite{mildenhall2021nerf} based methods involves meticulous tuning of the density threshold, leading to artifacts due to the inherent ambiguity in the density field, as previously pointed out in Unisurf \cite{oechsle2021unisurf}.
\\{\bf Generalizable Neural Surface Reconstruction.} 
Point Cloud input models \cite{peng2020convolutional,boulch2022poco,ouasfi2022few,ouasfi2024robustifying,ouasfi2024mixing} are typically trained with SDF ground truth supervision. 
To obtain 3D reconstructions from sparse images with an accurate level set, current methods marry generalizable NeRFs with the Signed Distance Function (SDF) based transformation functions that model the volume density, thereby enabling volume rendering. Among these methods, both SparseNeuS \cite{long2022sparseneus} and VolRecon \cite{ren2023volrecon} achieve generalizable neural surface reconstruction by utilizing information from source images as priors. SparseNeuS \cite{long2022sparseneus} does this by encoding geometric information using a regular euclidean volume, while VolRecon \cite{ren2023volrecon} introduces multi-view image features through the view transformer to advance this scheme. The most recent work of ReTR \cite{liang2024retr} utilizes a hybrid extractor to obtain a multi-level euclidean volume and then employs a reconstruction transformer to enhance performance. 
Contrarily to these methods, we employ the Unisurf \cite{oechsle2021unisurf} based volumetric rendering framework which reconstructs surfaces by predicting occupancy, enabling the seamless transfer of the opacity information from a pretrained GeoNeRF \cite{johari2022geonerf} to learn a refined occupancy field.

\section{Method}\label{sec:Method}

\begin{figure*}
\includegraphics[width=1.0\linewidth]{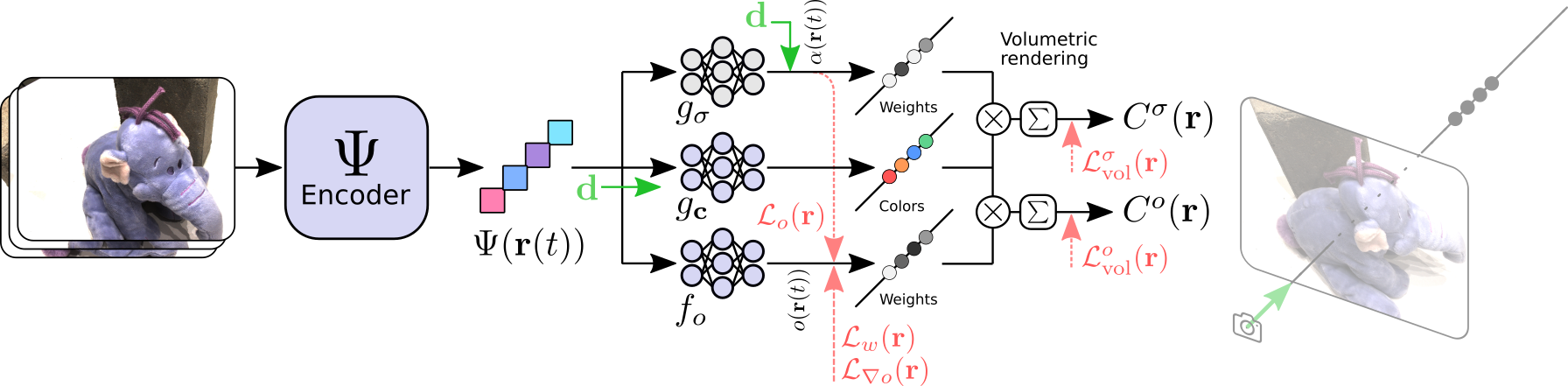}
\vspace{-7pt}
\caption{Overview of our transfer learning: Our final model (in light blue) is comprised of tuned image encoder $\Psi$ and implicit occupancy and color decoders $f_o$ and $g_{\bold{c}}$. The encoder $\Psi$, the color decoder $g_{\bold{c}}$ and density decoder $g_{\sigma}$ are initialized as a pretrained generalizable NeRF. Red dashed lines symbolize our tuning losses. We apply multiple regularizations on our occupancy $f_o$, while tuning the network with both the density and occupancy guided volumetric renderings.} 
\label{fig:pipe}
\vspace{-18pt}
\end{figure*}

Our goal is to obtain a feed-forward generalizable 3D reconstruction network from a sparse array of images. Namely, this network (\eg $f_{o}$) should be able to deliver an implicit shape representation, \ie a binary shape occupancy field for instance, from observed images $\{I_i\}_{i=1}^{N}$ and their calibrations $\{\pi_i\}_{i=1}^{N}$ of a test scene (unseen at training), without requiring any optimization on this new scene. The inferred shape $\hat{\mathcal{S}}$ can be obtained as the level set of the learned occupancy $f_{o}$:
\begin{equation}
\hat{\mathcal{S}} = \{\bold{x}\in\mathbb{R}^3 \mid f_{o}(\bold{x}) = 0.5\}.
\end{equation} 
Practically, an explicit triangle mesh for $\hat{\mathcal{S}}$ can be obtained through the Marching Cubes algorithm \cite{lorensen1998marching} while querying the neural network $f_{o}$. We propose to adapt an existing generalizable NeRF model for this purpose. We chose the model denoted GeoNeRF \cite{johari2022geonerf} without loss of generality as it is one of the best performing models in generalizable novel view synthesis. 
\vspace{-5pt}
\paragraph{\textbf{Generalizable NeRF model}}
Let $\bold{r}$ be a ray, \ie $\bold{r}(t) = \bold{o} + t\bold{d}$, where $\bold{o}$ is the camera origin and $\bold{d}$ the ray direction. The color $C$ of the pixel corresponding to a ray $\bold{r}$ can be generated by integrating along the ray:   
\begin{align} 
C(\bold{r}) &= \int T(t)\sigma(\bold{r}(t))\bold{c}(\bold{r}(t),\bold{d})dt\\
&= \int w(t) \bold{c}(\bold{r}(t),\bold{d})dt.
\label{equ:volrend}
\end{align}
In the equation above, $\sigma(\bold{r}(t))$ denotes volume density, which represents a differential opacity signaling the amount of radiance  accumulated by a ray passing through the point $\bold{r}(t)$. $T(t)$ denotes transparency, \ie the accumulated transmittance along the ray until $t$, which can be derived from density accordingly:  
\begin{equation}
T(t) = \exp\left(-\int^{t}\sigma(\bold{r}(s)) ds\right).
\label{equ:trans}
\end{equation}
Furthermore, $w(t)$ is commonly referred to as the volumetric rendering weighting function.

In practice, the integral in Equation \ref{equ:volrend} is approximated using discrete samples $\{t_i\}$ with the quadrature rule \cite{max1995optical}. This gives rise to an equation resembling $\alpha$-compositing, where $\alpha$ represents opacity, which controls the amount of radiance that is absorbed or transmitted at each point in the scene. This leads to :
\begin{align}
C_{\sigma}(\bold{r}) &= \sum_i T_i \left(1-e^{-\sigma_i(t_{i+1}-t_i)}\right)\bold{c}(\bold{r}(t_i),\bold{d})\\
&= \sum_i \prod_{j=0}^{i-1}(1-\alpha_j) \alpha_i \bold{c}(\bold{r}(t_i),\bold{d})
\label{equ:disc}
\end{align}
where $\alpha_i = 1-\exp(-\sigma_i(t_{i+1}-t_i))$. 

Gneralizable NeRFs are typically comprised of an encoder network (\eg $\Psi$) producing spatially local features. These features are mapped subsequently by a density network (MLP) (\eg $g_\sigma$) and a viewing direction dependent color network (MLP) (\eg $g_\bold{c}$).
(\cf Figure\ref{fig:pipe}). Inference is achieved via volumetric  rendering as shown above. The colors and densities necessary for this rendering are thus modelled as follows:
\begin{align} 
\bold{c}(\bold{r}(t),\bold{d}) &= g_\sigma(\Psi(\bold{r}(t),\{I_i\}),\bold{d}) \\ 
\sigma(\bold{r}(t)) &= g_\bold{c}(\Psi(\bold{r}(t),\{I_i\})).
\end{align}

\paragraph{\textbf{Adapting a Generalizable NeRF model}}

A geometry can be extracted from NeRF \cite{mildenhall2021nerf} models at test time by thresholding the density function, which is view and sampling independent. However, it is not clear how the threshold can be selected, and the obtained geometries tend to be noisy and inaccurate, with high Chamfer errors with respect to the ground truth. 

Although the opacity is the closest measure in a NeRF \cite{mildenhall2021nerf} to an occupancy, the opacity as defined in the volumetric rendering framework is view and ray-sampling dependent. Hence it is not clear how a solely spatially dependent geometry can be extracted from it.

Yet, in order for generalizable NeRFs to be able to reason correctly about 3D for accurate novel view synthesis, we hypothesize that they must encompass a good high level representation of geometry, that could be nudged towards an accurate and smooth shape representation per se. Based on this, we propose to tune a Generalizable NeRF (Namely GeoNeRF \cite{johari2022geonerf}) into a generalizable occupancy model, that offers view and ray-sampling independent geometry at convergence.

We define a new Sigmoid activated implicit decoder  $f_{o}$ that will represent the occupancy field in feature space:
\begin{equation}
o(\bold{r}(t))=f_{o}(\Psi(\bold{r}(t),\{I_i\})
\end{equation}

We propose to tune the encoder $\Psi$ to adapt the feature space to this new prediction task. This can be seen as form of transfer learning. As this tuning is achieved through volumetric rendering, we use the baseline model color network ($\bold{c}:=g_{\bold{c}}$) to perform volumetric rendering with $f_o$: 
\begin{equation}
C_{o}(\bold{r}) 
= \sum_i \prod_{j=0}^{i-1}(1-o_j) o_i \bold{c}(\bold{r}(t_i),\bold{d}).
\end{equation}

Color network $g_\bold{c}$ has been pretrained using the density based weight functions in volumetric rendering. Hence, it is important to tune it as-well to this new occupancy based rendering. Furthermore, as we also want out initial Generalizable NeRF not to deviate substantially from its original weights and hence lose its original knowledge, we train by backpropagating both the volumetric rendering loss based on occupancy, and the original volumetric rendering loss based on density together: 
\begin{align}
\mathcal{L}_{vol}^o(\bold{r}) &= ||C_o(\bold{r}) -  C_{\text{\tiny GT}}(\bold{r})||_2\\
\mathcal{L}_{vol}^{\sigma}(\bold{r}) &= ||C_\sigma(\bold{r}) -  C_{\text{\tiny GT}}(\bold{r})||_2.
\end{align}

We also retain the depth supervision used by the geometry reasoner in GeoNeRF \cite{johari2022geonerf}, with ground truth depths if available, and with their self-supervised depth loss otherwise. Based on the knowledge that assuming solid objects, $\alpha$ becomes a discrete occupancy indicator
variable $o \in \{0, 1\}$ which either takes $o = 0$ in free space and $o = 1$ in occupied space, we bootstrap our occupancy with $\alpha$ predictions from the density branch, as a form of warm up or initialization for a few iterations at the beginning of the training:
\begin{equation}
\mathcal{L}_o(\bold{r}) = \sum_i ||o_i - \alpha_i||_2. 
\end{equation}

A common problem with learning geometry through volumetric rendering is that the weight function does not peak at the surface \cite{oechsle2021unisurf, wang2021neus}. This can be observed in Figure \ref{fig:weight_dist}, where neither our baseline nor the generalizable NeRF model we build on satisfy this constraint. Hence, we propose a novel loss to remedy this limitation. First, for the current ray $\bold{r}$, we perform ray tracing, \ie finding sample pair $t_k$ and $t_{k+1}$ where the occupancy flips from "empty" ($f_o(t_k)< 0.5$) to "occupied" ($f_o(t_{k+1})\geq0.5$) for the first time along the ray. Then, we perform secant method based root finding \cite{niemeyer2020differentiable, oechsle2021unisurf} between these samples to find the root $t^*$ corresponding to the surface-ray intersection. Ideally, we want our weight function to reach a sharp $1$-peak at this location. Hence, we supervise the weight with a Gaussian centered at root $t^*$, and whose standard deviation we dynamically reduce during training, following the scheduling detailed in Section \ref{sec:Experiments}:       
\begin{equation}
\mathcal{L}_w(\bold{r}) = \sum_t ||\prod_{j=0}^{i-1}(1-o_j) o_i - \text{e}^{-((t_i-t^*)/a)^2} ||_2. 
\label{equ:weight_rendering}
\end{equation}

To reduce the noise in our reconstructions, we follow \cite{oechsle2021unisurf} and implement a smoothing over the normalized spatial gradients of our occupancy function at the surface, \ie using root locations $t^*$: 
\begin{equation}
\mathcal{L}_{\nabla o}(\bold{r}) = ||\frac{\nabla o(\bold{r(t^*)})}{||\nabla o(\bold{r(t^*)})||_2} - \frac{\nabla o(\bold{r(t^*)+\epsilon})}{||\nabla o(\bold{r(t^*)+\epsilon})||_2}||_2, 
\end{equation}
where $\epsilon \in \mathbb{R}^3$ is a small random perturbation, and gradients can be computed efficiently through automatic differentiation \cite{paszke2019pytorch}. 

Finally, our full training is done following the combined objective averaged over batches of rays, and we train on the same multi-view data as our original Generalizable NeRF model was initially trained on: 
\begin{equation}
\mathcal{L} = \sum_{\bold{r}} \mathcal{L}_{vol}^o(\bold{r}) + \mathcal{L}_{vol}^\sigma(\bold{r}) + \lambda\mathcal{L}_{\nabla o}(\bold{r}) + \mu\mathcal{L}_w(\bold{r}) + \nu\mathcal{L}_o(\bold{r}).
\end{equation}
These loses are analysed separately in section \ref{sec:Experiments}, where we show their respective numerical and qualitative contribution to our overall performance.

\begin{table*}[t]
\scalebox{0.9}{
\centering
\hspace*{-\leftmargin}\begin{tabular}{llllllllllllllll|l}
\hline
\textbf{Scan} & 24   & 37   & 40   & 55   & 63   & 65   & 69   & 83   & 97   & 105  & 106  & 110  & 114  & 118  & 122  & Mean \\  
\hline
\hline
COLMAP \cite{schonberger2016structure} & \textbf{0.9} & 2.89 & 1.63 & 1.08 & 2.18 & 1.94 & 1.61 & 1.3  & 2.34 & 1.28 & 1.1  & 1.42 & 0.76 & 1.17 & 1.14 & 1.52 \\  
MVSNet \cite{yao2018mvsnet} & 1.05 & 2.52 & 1.71 & 1.04 & 1.45 & \textbf{1.52} & 0.88 & 1.29 & 1.38 & 1.05 & 0.91 & \underline{0.66} & 0.61 & 1.08 & 1.16 & 1.22 \\  
\hline
IDR \cite{yariv2020multiview} & 4.01 & 6.4  & 3.52 & 1.91 & 3.96 & 2.36 & 4.85 & 1.62 & 6.37 & 5.97 & 1.23 & 4.73 & 0.91 & 1.72 & 1.26 & 3.39 \\  
VolSDF \cite{yariv2021volume} & 4.03 & 4.21 & 6.12 & \underline{0.91} & 8.24 & 1.73 & 2.74 & 1.82 & 5.14 & 3.09 & 2.08 & 4.81 & 0.6  & 3.51 & 2.18 & 3.41 \\  
UNISURF \cite{oechsle2021unisurf} & 5.08 & 7.18 & 3.96 & 5.3  & 4.61 & 2.24 & 3.94 & 3.14 & 5.63 & 3.4  & 5.09 & 6.38 & 2.98 & 4.05 & 2.81 & 4.39 \\  
NeuS \cite{wang2021neus} & 4.57 & 4.49 & 3.97 & 4.32 & 4.63 & 1.95 & 4.68 & 3.83 & 4.15 & 2.5  & 1.52 & 6.47 & 1.26 & 5.57 & 6.11 & 4.00 \\  
IBRNet-ft \cite{wang2021ibrnet} & 1.67 & 2.97 & 2.26 & 1.56 & 2.52 & 2.30 & 1.50 & 2.05 & 2.02 & 1.73 & 1.66 & 1.63 & 1.17 & 1.84 & 1.61 & 1.90 \\  
SparseNeuS-ft \cite{long2022sparseneus} & 1.29 & 2.27 & 1.57 & \textbf{0.88} & 1.61 & 1.86 & 1.06 & \underline{1.27} & 1.42 & 1.07 & 0.99 & 0.87 & \textbf{0.54} & 1.15 & 1.18 & 1.27 \\  
\hline
PixelNerf \cite{yu2021pixelnerf} & 5.13 & 8.07 & 5.85 & 4.4  & 7.11 & 4.64 & 5.68 & 6.76 & 9.05 & 6.11 & 3.95 & 5.92 & 6.26 & 6.89 & 6.93 & 6.28 \\ 
IBRNet \cite{wang2021ibrnet} & 2.29 & 3.70 & 2.66 & 1.83 & 3.02 & 2.83 & 1.77 & 2.28 & 2.73 & 1.96 & 1.87 & 2.13 & 1.58 & 2.05 & 2.09 & 2.32 \\  
MVSNeRF \cite{chen2021mvsnerf} & 1.96 & 3.27 & 2.54 & 1.93 & 2.57 & 2.71 & 1.82 & 1.72 & 2.29 & 1.75 & 1.72 & 1.47 & 1.29 & 2.09 & 2.26 & 2.09 \\ 
GeoNeRF \cite{johari2022geonerf} & 3.40 & 4.37 & 3.99 & 2.94 & 5.08 & 4.50 & 3.42 & 4.68 & 4.54 & 4.05 & 3.47 & 3.23 & 3.34 & 3.57 & 3.63 & 3.88 \\ 
SparseNeuS \cite{long2022sparseneus} & 1.68 & 3.06 & 2.25 & 1.1  & 2.37 & 2.18 & 1.28 & 1.47 & 1.8  & 1.23 & 1.19 & 1.17 & 0.75 & 1.56 & 1.55 & 1.64 \\ 
VolRecon \cite{ren2023volrecon} & 1.2  & 2.59 & 1.56 & 1.08 & 1.43 & 1.92 & 1.11 & 1.48 & 1.42 & 1.05 & 1.19 & 1.38 & 0.74 & 1.23 & 1.27 & 1.38 \\  
ReTR \cite{liang2024retr} & 1.05 & 2.31 & \textbf{1.44} & 0.98 & \textbf{1.18} & \textbf{1.52} & 0.88 & 1.35 & \underline{1.3}  & \textbf{0.87} & 1.07 & 0.77 & 0.59 & 1.05 & 1.12 & 1.17 \\
\hline
Ours & 1.01 & \underline{2.24} & 1.52 & \textbf{0.88} & 1.37 & 1.82 & \underline{0.85} & 1.39 & \textbf{1.25} & \underline{1.0} & \textbf{0.77} & \underline{0.63} & \underline{0.57} & \underline{0.96} & \textbf{1.0} & \underline{1.15} \\  
$Ours^{\ast}$ & \underline{0.95} & \textbf{2.23} & \underline{1.45} & 0.94 & \underline{1.26} & \underline{1.67} & \textbf{0.81} & \textbf{1.21} & 1.34 & 1.02 & \underline{0.84} & \textbf{0.6} & 0.58 & \textbf{0.94} & \underline{1.02} & \textbf{1.12} \\ 
\hline
\end{tabular}
}
\vspace{10pt}
\caption{Quantitative comparison on the DTU dataset \cite{aanaes2016large}. Best and second best methods are \textbf{emboldened} and \underline{underlined} respectively. $Ours^{\ast}$ refers to the model trained using additional datasets.} \label{tab:table1}
\vspace{-30pt}
\end{table*}

\vspace{-2pt}
\section{Experiments}\label{sec:Experiments}

\begin{figure*}
    \centering
    \includegraphics[width=1.0\linewidth]{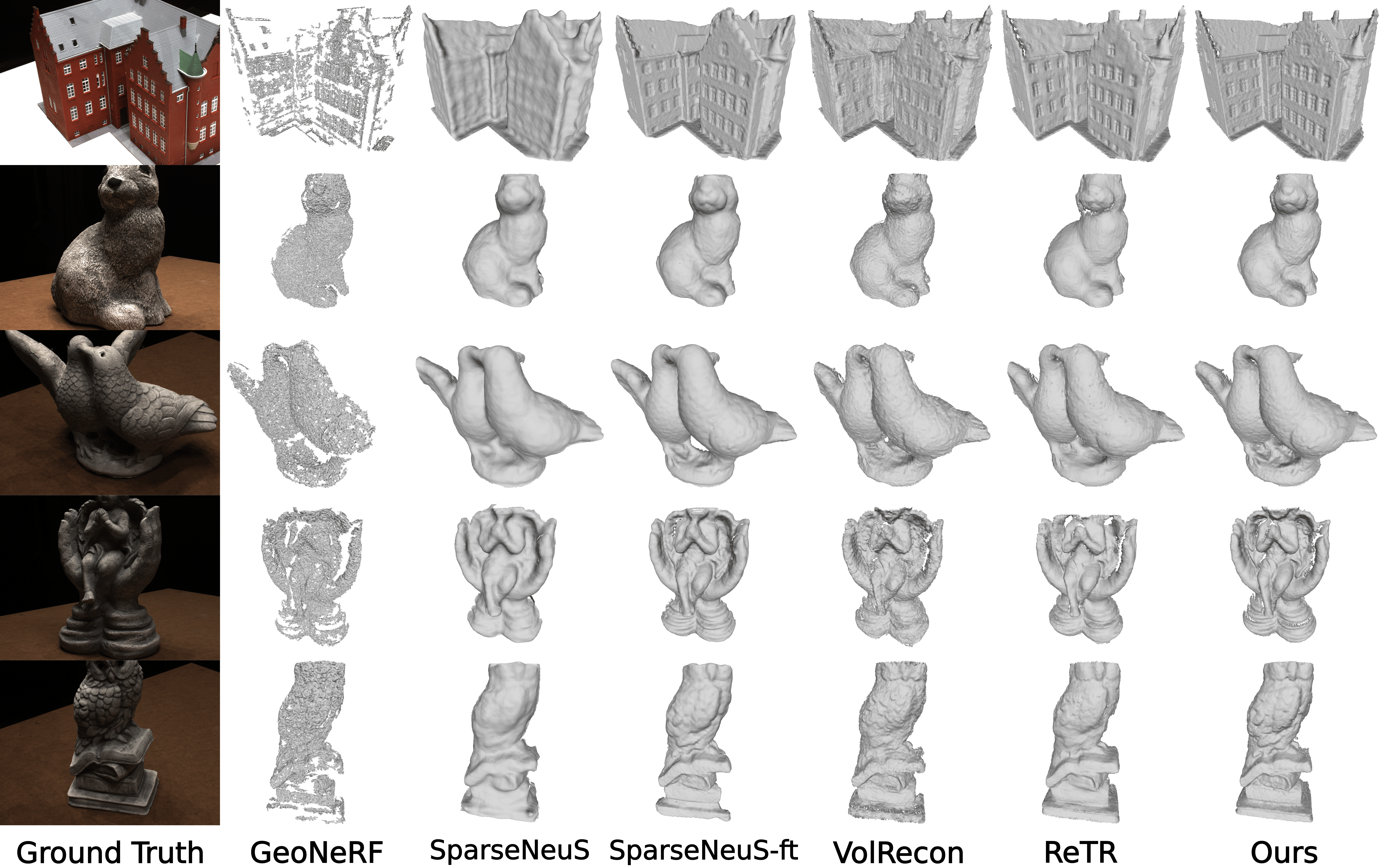}
    \vspace{-10pt}
    \caption{Qualitative comparison of reconstructions from 3 input views in datatset DTU.}
    \label{fig:DTU}
\end{figure*}

\begin{figure*}
    \centering
    \includegraphics[width=0.9\linewidth]{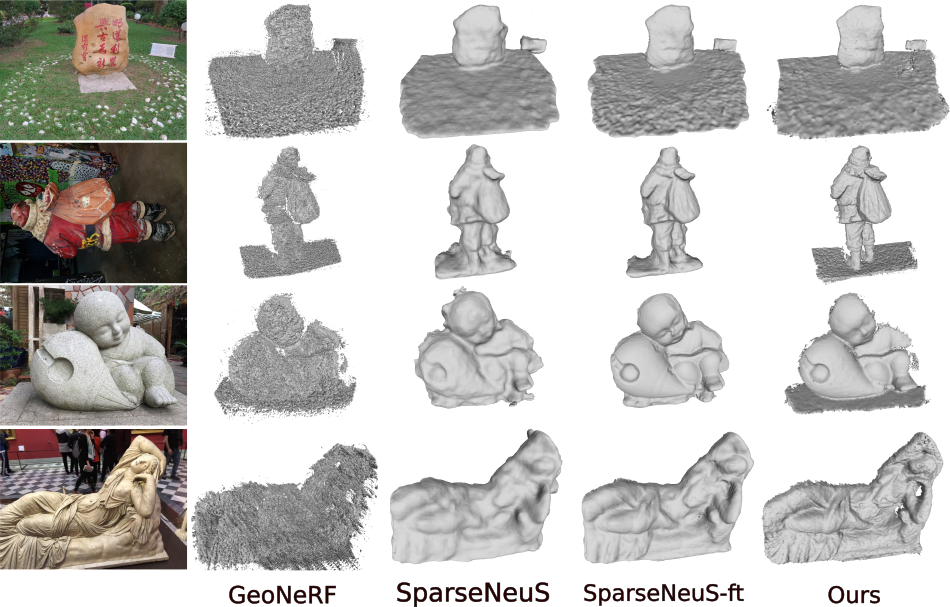}
    \vspace{-10pt}
    \caption{Qualitative comparison of reconstructions from 3 input views in datatset BMVS. Note that we reconstruct detailed surfaces with our method without any fine-tuning.}
    \label{fig:BMVS}
\end{figure*}

In this section, we showcase the efficacy and merits of our proposed approach. Firstly, we offer an intricate overview of our experimental configurations, encompassing implementation specifics, datasets, and baseline methods. Secondly, we present both quantitative and qualitative comparisons on two extensively utilized datasets, namely DTU \cite{aanaes2016large} and BlendedMVS \cite{yao2020blendedmvs}. Finally, we perform thorough ablation studies to scrutinize the impact of distinct components in our proposed methodology.
\vspace{-7pt}
\paragraph{\textbf{Datasets}}
In line with prior research \cite{long2022sparseneus, ren2023volrecon, liang2024retr}, we employ the DTU dataset \cite{aanaes2016large} for the training phase. Since we perform transfer learning from GeoNeRF \cite{aanaes2016large}, which is a NeRF-based framework, to learn generalizable occupancy fields, we are also able to leverage training on non-object centric datasets such as the real forward-facing datasets from LLFF \cite{mildenhall2019local} and IBRNet \cite{wang2021ibrnet} and hence, we also include an additional model to show the effect of training on more data. In both cases (with and without additional training datasets), our backbone GeoNeRF \cite{aanaes2016large} was trained on the same data as our full models to maintain fairness in evaluation performance comparisons. The DTU dataset \cite{aanaes2016large} is characterized by indoor multi-view stereo data, featuring ground truth point clouds from 124 distinct scenes and under 7 different lighting conditions. Throughout our experiments, we utilize the same set of 15 scenes as \cite{long2022sparseneus, ren2023volrecon, liang2024retr} for testing purposes, reserving the remaining scenes for training. Concerning the BlendedMVS dataset \cite{yao2020blendedmvs}, we opt for 7 scenes in accordance with SparseNeuS \cite{long2022sparseneus}. For each scene, we use the same $3$ sparse input views following SparseNeuS \cite{long2022sparseneus}. To ensure impartial evaluation, we use the foreground masks from IDR \cite{yariv2020multiview} to evaluate how well our approach performs on the test set, consistent with prior research \cite{long2022sparseneus, ren2023volrecon, liang2024retr}. Additionally, to assess the generalization capability of our proposed framework, we qualitatively compare our method on the BlendedMVS dataset \cite{yao2020blendedmvs} without any fine-tuning. For our novel-view synthesis experiments, we follow the same split of testing images within a scene as in GeoNeRF \cite{aanaes2016large} and the testing scenes are identical to our 3D reconstruction experiments.
\vspace{-7pt}
\paragraph{\textbf{Baselines}}
In order to showcase the efficacy of our method, we conducted comparisons with a) SparseNeus \cite{long2022sparseneus}, VolRecon \cite{ren2023volrecon} and ReTR \cite{liang2024retr}, the leading generalizable neural surface reconstruction approaches and their finetuned (ft) versions on sparse images; b) Generalizable neural rendering methods \cite{yu2021pixelnerf, wang2021ibrnet, chen2021mvsnerf, johari2022geonerf}; c) Neural implicit reconstruction methods \cite{yariv2020multiview, yariv2021volume, oechsle2021unisurf, wang2021neus} that necessitate scene-specific training from the beginning; and finally, d) Well-known multi-view stereo (MVS) \cite{schonberger2016structure, yao2018mvsnet} methods. 
\vspace{-7pt}
\paragraph{\textbf{Implementation details}}
Our model is implemented using PyTorch \cite{paszke2019pytorch} and PyTorch Lightning \cite{falcon2019pytorch}. During the training phase, we utilize an image resolution of $800 \times 600$, with $N$ (the number of source images) set to $3$. Training extends over $5400$ steps using the Adam optimizer \cite{kingma2014adam} on a single RTX A$6000$ GPU, with an initial learning rate of $5 \times 10^{-4}$. We apply the occupancy distillation loss for the first $100$ warm-up steps. Thereafter, the weight of the distillation loss is reduced to $0$. A cosine learning rate scheduler \cite{loshchilov2016sgdr} is applied to the optimizer. The ray number sampled per batch and the batch size are configured to $128$ and $1$, respectively. Employing a hierarchical sampling strategy, we uniformly sample $N_{coarse}$ points on the ray during both training and testing. Subsequently, importance sampling is applied to sample an additional $N_{fine}$ points on top of the coarse probability estimation. In our experiments, we set $N_{coarse}$ to $96$ and $N_{fine}$ to $32$. During testing, the image resolution is kept the same at $800 \times 600$. We also address the scheduling strategy followed for $\mathcal{L}_{w}$, described in \ref{equ:weight_rendering}. Specifically, we follow :
\begin{equation}
a = \max(a_{max}\text{e}^{-\text{k}\beta}, a_{min}). 
\end{equation}
where $k$ denotes the global iteration number. $a_{max}$, $a_{min}$ and $\beta$ are hyperparameters, which we set to $1$, $0.04$ and $0.001$ respectively. Finally, the weights associated with the losses are set as $\lambda = 0.1$, $\mu = 0.2$ and $\nu = 0.2$. We use the Marching Cubes algorithm \cite{lorensen1998marching} with a grid resolution of $400$ for extracting each surface mesh by thresholding the occupancy field at $0.5$.

\subsection{Sparse View Reconstruction on DTU}
We conduct surface reconstruction with sparse views (only $3$ views) on the DTU dataset \cite{aanaes2016large} and assess the predicted surface by comparing it to the ground-truth point clouds using the chamfer distance metric. To facilitate a fair comparison, we followed the evaluation process employed in \cite{long2022sparseneus, ren2023volrecon, liang2024retr} and adhered to the same testing split as described in them. As indicated in Table \ref{tab:table1}, our method (only DTU trained) surpasses SparseNeuS \cite{long2022sparseneus} and VolRecon \cite{ren2023volrecon} by a considerable margin, \ie by $30\%$ and $17\%$ respectively. When we use additional datasets for training, the gap further increases to $32\%$ and $19\%$ respectively. Furthermore, our approach exhibits superior performance compared to well-known multi-view stereo (MVS) methods like Colmap \cite{schonberger2016structure} and MVSNet \cite{yao2018mvsnet}. Our approach also demonstrates superior performance in comparison to ReTR \cite{liang2024retr}, which is the latest state-of-the-art generalizable neural implicit reconstruction method. Additionally, we present qualitative results of sparse view reconstruction in Fig. \ref{fig:DTU}, showcasing that our reconstructed geometry features more expressive and detailed surfaces which represent the ground truth surfaces more accurately when compared to the current state-of-the-art methods. Another DTU generalization experiment using MVSNeRF \cite{chen2021mvsnerf} as the backbone is included in the supplemetary section to illustrate that the method presented can be extended to other generalizable NeRF baselines.
\vspace{-5pt}
\subsection{Generalization on BlendedMVS}
To demonstrate the generalization prowess of our proposed approach, we perform additional evaluations on the BlendedMVS dataset \cite{yao2020blendedmvs} without resorting to any fine-tuning. The high-fidelity reconstructions of large-scale scenes and small objects across diverse domains, as illustrated in Fig. \ref{fig:BMVS}, affirms the efficacy of our method in terms of its generalization capabilities. Our method is able to obtain more detailed surfaces in comparison to SparseNeuS \cite{long2022sparseneus}, even after it is fine-tuned on the sparse testing set.
\vspace{-5pt}

\subsection{Novel-view synthesis performance on DTU}
Since GeoNeRF \cite{johari2022geonerf} is among the state-of-the-art methods for sparse novel-view synthesis, it is reasonable to assume that we also inherit its novel-view synthesis performance since we train our method by transferring its features. To validate this, we evaluate novel-view synthesis results on the DTU \cite{aanaes2016large} dataset. We compare our method against GeoNeRF \cite{johari2022geonerf}, VolRecon \cite{ren2023volrecon} and ReTR \cite{liang2024retr} on the mean PSNR, SSIM and LPIPS metrics over all scenes. Since we use $3$ input source images to build our cost feature volume, for fairness in comparison, we also evaluate the novel-view synthesis results for VolRecon \cite{ren2023volrecon} and ReTR \cite{liang2024retr} with $3$ input source images to build the cost feature volume, instead of the $4$ input source images used during their respective trainings. Our results can be summarized as below :

\begin{table}[h]
\centering
{
\hspace*{1.2\leftmargin}\begin{tabular}{ c|ccc}
 \hline
 Method & PSNR $\uparrow$ &SSIM $\uparrow$ &LPIPS $\downarrow$\\
 \hline
 VolRecon \cite{ren2023volrecon} & 19.61 & 0.81 & 0.23\\
 ReTR \cite{liang2024retr} & 19.53 & 0.79 & 0.24\\
 GeoNeRF \cite{johari2022geonerf} & 23.48 & \textbf{0.93} & \textbf{0.087}\\
 Ours & \textbf{24.08} & \textbf{0.93} & 0.093\\
 \hline
\end{tabular}
}
\vspace{5pt}
\caption{Novel-view performance on DTU \cite{aanaes2016large}.} \label{tab:novel_view}
\vspace{-25pt}
\end{table}

\begin{figure*}[t!]
    \centering
    \includegraphics[width=1.025\linewidth]{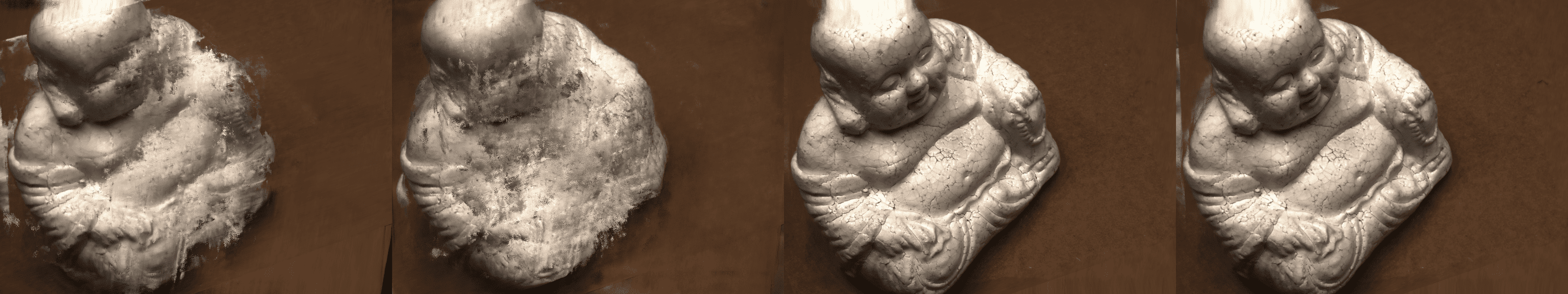} \\ 
    \includegraphics[width=1.025\linewidth]{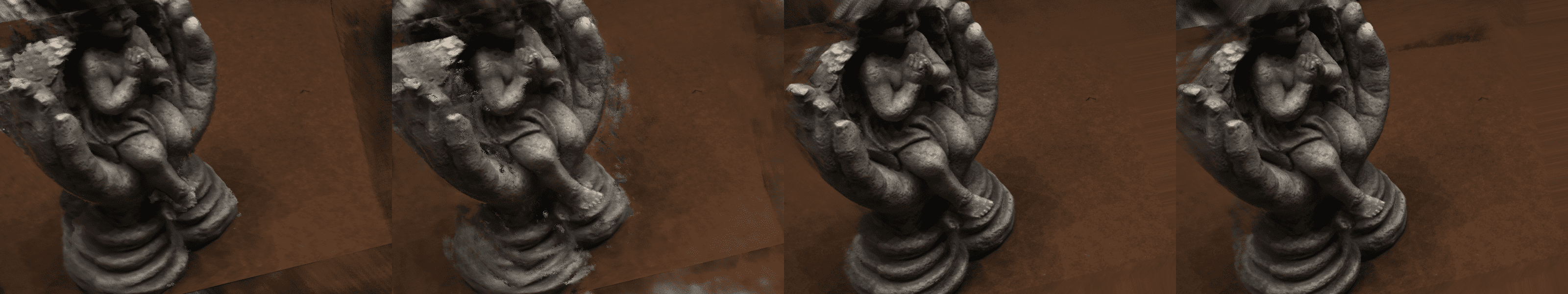} \\ 
    \includegraphics[width=1.025\linewidth]{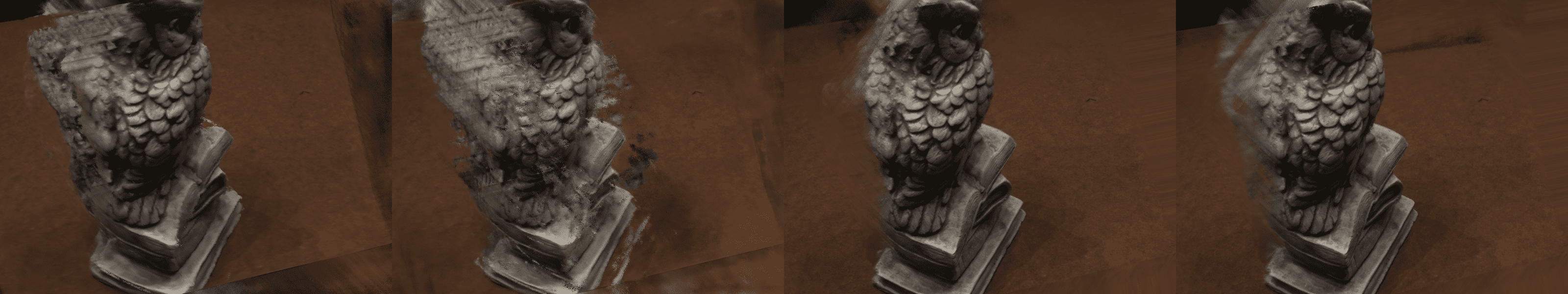} \\ 
    \includegraphics[width=0.825\linewidth]{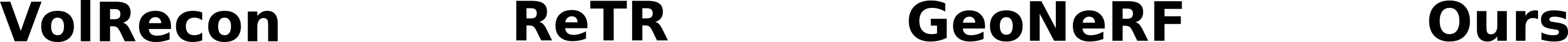} \\ 
    \caption{Novel-View synthesis qualitative evalutation on DTU \cite{aanaes2016large} using $3$ source images. We notice that generalizable reconstruction models (ReTR \cite{liang2024retr}, VolRecon \cite{ren2023volrecon}) struggle to perform extreme novel-view extrapolation.}
    \vspace{-15pt}
    \label{tab:nvs_images3}
\end{figure*}

As indicated in Table \ref{tab:novel_view}, we are very close in performance on novel-view synthesis metrics to GeoNeRF \cite{johari2022geonerf}. Note that we use $3$ input views, the same setting on which we conduct our novel-view experiments on. The original GeoNeRF \cite{johari2022geonerf}, however, was trained with $6$ input views. The purpose behind this experiment was simply to make sure that our training paradigm does not sacrifice GeoNeRF \cite{johari2022geonerf}'s excellent novel-view synthesis performance. Overall, our method offers state-of-the-art reconstruction performance without foregoing the novel-view synthesis performance of one of the state-of-the-art novel-view synthesis methods we transfer features from. This is because our learnt occupancy fields by design and by penalization with the help of occupancy loss $\mathcal{L}_{o}$ induce volumetric weights associated with the opacity fields of GeoNeRF \cite{johari2022geonerf}. We also notice that we far surpass VolRecon \cite{ren2023volrecon} and ReTR \cite{liang2024retr} in terms of novel-view synthesis performance. Hence, we offer a dual advantage, leading to our method offering \textbf{state-of-the-art} reconstruction results along with superior novel-view synthesis results. We also show some qualitative comparisons in Figure \ref{tab:nvs_images3} and several other qualitative comparisons on several DTU \cite{aanaes2016large} scenes in the supplementary section.

\vspace{-10pt}
\subsection{Ablation studies}
In this section, we conduct an ablative analysis to justify the choice of our final architecture and choice of losses. We ablate in the full training scenario using all testing scenes of the DTU dataset \cite{aanaes2016large}. 

\vspace{-7pt}
\begin{table}[h!]
\centering
\vspace{-5pt}
\begin{minipage}{0.45\linewidth}
\centering
\scalebox{1.0}{
\begin{tabular}{ c|c}
 \hline
 Method & Chamfer $\downarrow$\\
 \hline
 w/o tuning encoder &1.29\\
 w tuning encoder &1.15\\
 \hline
\end{tabular}
}
\caption{Effect of fine-tuning encoder/feature vol. of GeoNeRF \cite{johari2022geonerf}} \label{tab:table2}
\end{minipage}
\hfill
\begin{minipage}{0.45\linewidth}
\centering
\scalebox{1.0}{
\begin{tabular}{ c|cc}
 \hline
 & w/o $\mathcal{L}_{\nabla o}$ &w $\mathcal{L}_{\nabla o}$\\
 \hline
 NC. $\uparrow$ & 0.64 & 0.68\\
 \hline
\end{tabular}
}
\caption{Normal Consistency} \label{tab:table6}
\end{minipage}
\vspace{-25pt}
\end{table}

\paragraph{\textbf{Effect of fine-tuning strategies}} In this study, we showcase the outcomes of training our occupancy model while fine-tuning different components of GeoNeRF \cite{johari2022geonerf}. As illustrated in Table \ref{tab:table2}, a better occupancy field is learnt when the encoder of the pre-trained GeoNeRF \cite{johari2022geonerf} model is tuned jointly with the occupancy network. This is because without doing so, the encoder is more suited for learning a density field as is the case for GeoNeRF \cite{johari2022geonerf}, while we aim to learn an occupancy field.

\vspace{-5pt}
\paragraph{\textbf{Effect of Different Loss Components}} In this study, we showcase the outcomes of various loss components to illustrate their efficacy. As illustrated in the $1^{st}$ row of Table \ref{tab:table4}, the novel volumetric rendering weight loss $\mathcal{L}_{w}$ is vital in learning a sharp and accurate occupancy field. Without it, the chamfer metric degrades, which demonstrates its importance. The $2^{nd}$ row illustrates the importance of bootstrapping our occupancy with $\alpha$ predictions. Furthermore, we see in Table \ref{tab:table3} that the accuracy chamfer metric expectedly suffers more without $\mathcal{L}_{w}$ as it is directly responsible for sharper surfaces, and less so for the completeness metric. 

\begin{table}[h!]
\vspace{-5pt}
\begin{minipage}{0.45\linewidth}
\centering
\scalebox{1}{
\begin{tabular}{ c|c}
 \hline
 Method & Chamfer $\downarrow$\\
 \hline
 w/o $\mathcal{L}_{w}$ &1.22\\
 w/o $\mathcal{L}_{o}$ &1.17\\
 \textbf{Full model} &1.15\\
 \hline
\end{tabular}
}
\caption{Effect of discarding different loss components} \label{tab:table4}
\end{minipage}
$\quad$
\begin{minipage}{0.45\linewidth}
\centering
\scalebox{1}{
\begin{tabular}{ c|cc}
 \hline
 & w/o $\mathcal{L}_{w}$ &w $\mathcal{L}_{w}$\\
 \hline
 Acc. $\downarrow$ & 0.77 & 0.67\\
 Comp. $\downarrow$ & 1.67 & 1.64\\
 Overall. $\downarrow$ & 1.22 & 1.15\\
 \hline
\end{tabular}
}
\caption{Ablation of accuracy \& completeness} \label{tab:table3}
\end{minipage}
\vspace{-25pt}
\end{table}

\begin{figure}[h!]
\begin{minipage}{0.45\linewidth}
\centering
\scalebox{1}{
     \rotatebox{90}{\hspace{18pt} \tiny Ours \hspace{22pt} Ours (w/o $\mathcal{L}_{w}$) \hspace{12pt} Geonerf \cite{johari2022geonerf}}
     \begin{subfigure}[b]{0.52\linewidth}
         \includegraphics[width=\linewidth]{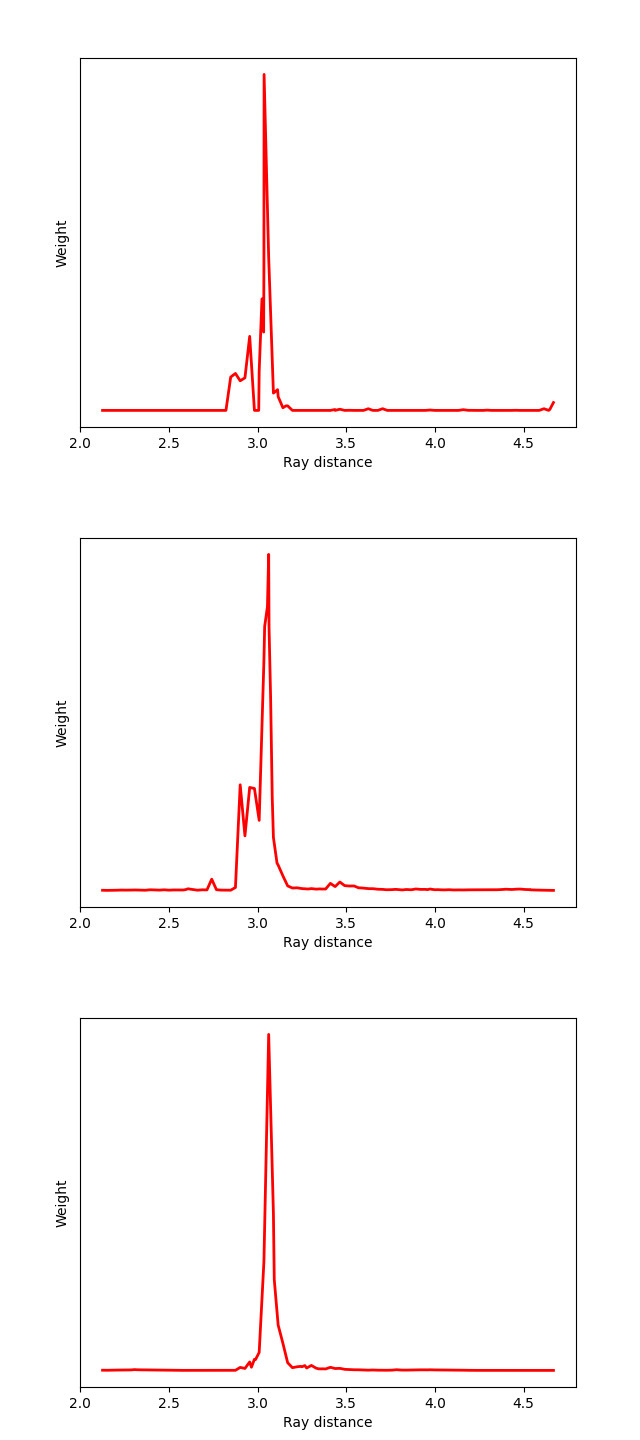}
     \end{subfigure}
     \begin{subfigure}[b]{0.52\linewidth}
         \includegraphics[width=\linewidth]{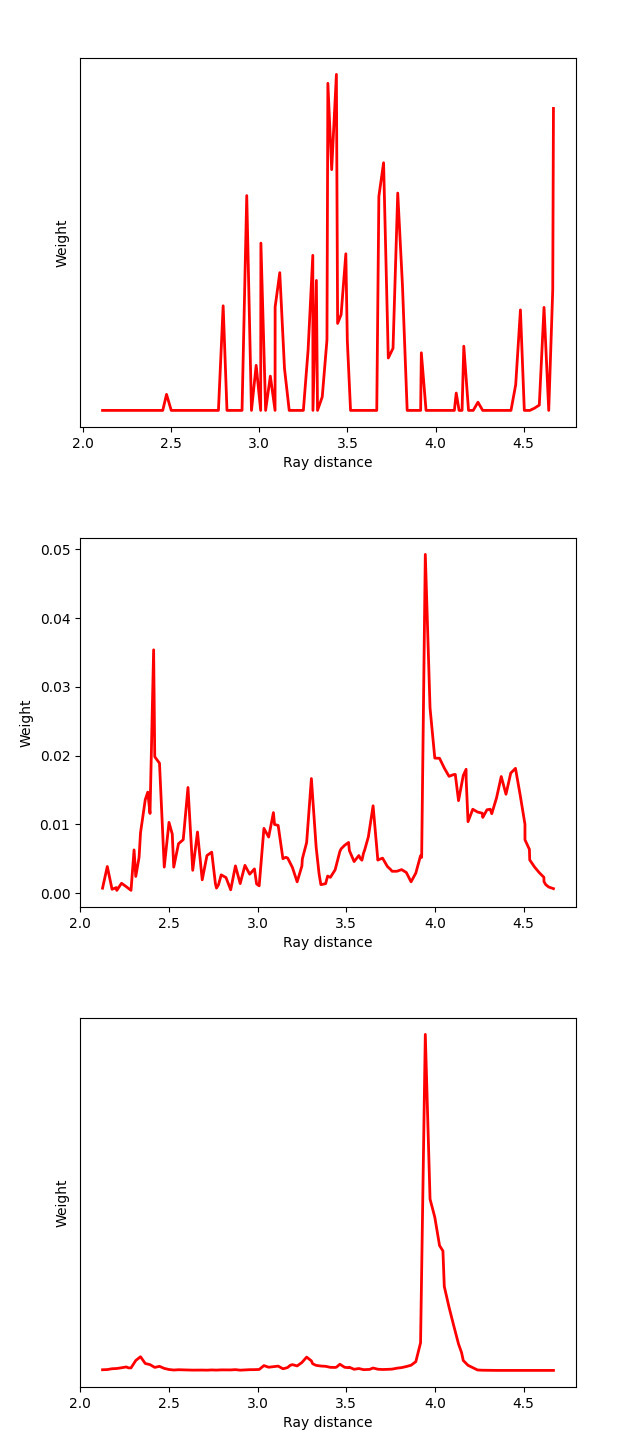}
     \end{subfigure}
}
 \caption{Comparison between volumetric weight distribution along rays between Geonerf \cite{johari2022geonerf} and our method.} \label{fig:weight_dist}
\end{minipage}
$\qquad$
\begin{minipage}{0.45\linewidth}
\centering
\begin{tabular}{ c|c}
 \hline
 Method & Training duration\\
 \hline
 SparseNeuS \cite{long2022sparseneus} & $\sim$3 days\\
 ReTR \cite{liang2024retr} & $\sim$3 days\\
 Ours (re-train) & \textbf{$\sim$3.5 hrs}\\
 \hline
\end{tabular}
 \caption{Comparison of total training duration.} \label{tab:novel_view2}
\end{minipage}
\vspace{-5pt}
\end{figure}

This is also illustrated in Fig.\ref{fig:weight_dist}, which shows that the loss results in a weight distribution that agrees with one theoretically induced by an occupancy field \ie one that $1$-peaks on the surface boundary, as discussed in Unisurf \cite{oechsle2021unisurf}. We also see in the $2^{nd}$ row that without our bootstrapped occupancy loss $\mathcal{L}_{o}$, our surface reconstruction performance is impacted negatively. Finally, we investigate the effect of discarding the normal loss $\mathcal{L}_{\nabla o}$. The normal loss is important for reducing noise in our reconstruction as witnessed in our qualitative results. This smoothing, can however, impact chamfer distance and it can be challenging to strike a perfect balance between these conflicting objectives. To illustrate the importance of normal loss, we provide normal consistency results in Tab.\ref{tab:table6} which demonstrates that our normal loss $\mathcal{L}_{\nabla o}$ smoothens the reconstructions and consequently, improves normal consistency between the reconstructed and GT meshes. The whole study demonstrates that all our losses are indispensable to our model's final performance. 

\paragraph{\textbf{Training time}} Here, we address the time of training for our model in Table \ref{tab:novel_view2}. Given a fully-trained GeoNeRF \cite{johari2022geonerf}, our model is able to finetune its features rapidly. While training SparseNeuS \cite{long2022sparseneus} and ReTR \cite{liang2024retr} fully on a single RTX A$6000$ GPU takes almost $3$ days, we are able to train to adapt features of a pretrained GeoNeRF \cite{johari2022geonerf} in approximately $3.5$ hrs to learn occupancy fields.
\vspace{-5pt}

\section{Limitations}
\vspace{-2pt}
Because of the underlying assumption while learning occupancy in a volumetric framework, our method is specifically designed to represent surfaces that are solid and not transparent. Moreover, the accuracy of the reconstructions is diminished in areas that are infrequently visible in the input images.

\vspace{-5pt}
\section{Conclusions}
\vspace{-2pt}
We introduce a novel approach, GeoTransfer, that rapidly transfers the 3D understanding of generalizable NeRFs to obtain accurate occupancy fields for implicit surface reconstruction. Unlike previous methods that introduced improvements to the encoders of generalizable SDF based reconstruction methods, we discovered that it sufficed to learn a occupancy network in feature space that learns to transform sampling-dependent opacities obtained from the state-of-the-art generalizable NeRF to sampling-independent occupancy fields. This approach also proved to be way faster than training a generalizable sparse 3D surface reconstruction method from scratch. Our method achieves state-of-the-art reconstruction quality for sparse inputs, showcasing its efficacy. 
\clearpage  

\title{GeoTransfer: Generalizable Few-Shot Multi-View Reconstruction via Transfer Learning\\– Supplementary Material –} 

\titlerunning{GeoTransfer}

\author{Shubhendu Jena \and
Franck Multon \and
Adnane Boukhayma}

\authorrunning{S.~Jena et al.}

\institute{Inria, Univ. Rennes, CNRS, IRISA, M2S, France}

\maketitle

We begin by applying our approach to another baseline model, specifically MVSNeRF \cite{chen2021mvsnerf}, and demonstrate significant improvements in its 3D surface reconstruction performance, both quantitatively and qualitatively. Thereafter, we provide additional ablation studies to justify our hyperparameter choices and also show qualitative video comparisons of our reconstruction results to other methods to visually demonstrate the impact of our approach and the corresponding losses. Qualitative comparisons with VolRecon \cite{ren2023volrecon} and ReTR \cite{liang2024retr} for our novel view synthesis results follow, and finally we conclude with some additional experimental details on our evaluation datasets of DTU \cite{aanaes2016large} and BlendedMVS \cite{yao2020blendedmvs}.

\section{Using MVSNeRF as backbone}
To demonstrate the generalizability of our approach, we apply it with MVSNeRF \cite{chen2021mvsnerf} as our backbone and denote the resulting model MVSTransfer. Tuning MVSNeRF \cite{chen2021mvsnerf} to MVSTransfer took us only $\sim$3 hrs on a single RTX A$6000$ GPU. Notice that training MVSNeRF \cite{chen2021mvsnerf} on the same GPU takes at least $3$ days. In the following comparative analysis, we are working with the same DTU \cite{aanaes2016large} reconstruction split trained MVSNeRF \cite{chen2021mvsnerf} to ensure fairness. Then, we use the same Sigmoid activated implicit decoder $f_{o}$ as in our experiments based in GeoNeRF \cite{johari2022geonerf}, and we leverage the same loss functions to obtain : 
\begin{equation}
\mathcal{L} = \sum_{\bold{r}} \mathcal{L}_{vol}^o(\bold{r}) + \mathcal{L}_{vol}^\sigma(\bold{r}) + \lambda\mathcal{L}_{\nabla o}(\bold{r}) + \mu\mathcal{L}_w(\bold{r}) + \nu\mathcal{L}_o(\bold{r}).
\end{equation}
The losses are weighted exactly as in the original paper and we finetune the decoder for $5400$ iterations to obtain a generalizable occupancy network. After running Marching Cubes algorithm \cite{lorensen1998marching} with a grid resolution of $400$ for extracting each surface mesh by thresholding the occupancy field at $0.5$ and computing chamfer metrics with respect to the ground-truth point clouds, we get :


\begin{table*}[h!]
\centering
\scalebox{0.96}{
\centering
\hspace*{-\leftmargin}\begin{tabular}{l|l|l|l|l|l|l|l|l|l|l|l|l|l|l|l}
\hline
\textbf{Scan} & 24   & 37   & 40   & 55   & 63   & 65   & 69   & 83   & 97   & 105  & 106  & 110  & 114  & 118  & 122 \\  
\hline
\hline
MVSNeRF \cite{chen2021mvsnerf} & 2.29 & \textbf{3.76} & 2.84 & 1.93 & 2.79 & 2.73 & 1.91 & 2.51 & 2.56 & 2.06 & 2.01 & 1.56 & 1.55 & 2.24 & 2.38 \\ 
MVSTransfer & \textbf{1.83} & 3.79 & \textbf{2.5} & \textbf{1.6} & \textbf{2.12} & \textbf{2.51} & \textbf{1.5} & \textbf{1.99} & \textbf{1.99} & \textbf{1.55} & \textbf{1.51} & \textbf{1.53} & \textbf{1.16} & \textbf{1.65} & \textbf{2.0} \\  
\hline
\end{tabular}
}
\vspace{10pt}
\caption{Quantitative comparison on the DTU dataset \cite{aanaes2016large}. The best method is \textbf{emboldened}.} \label{tab:mvsnerf}
\end{table*}
Qualitatively, our approach leads to a globally more accurate learning of the occupancy fields, leading to the extracted mesh to be wrapped more closely around the ground truth meshes (obtained by running screened poisson surface reconstruction on the ground truth point clouds). Some examples are shown in Figure \ref{tab:mvsnerf_comparison} as follows : 

\begin{figure*}[h!]
    \centering
    \includegraphics[width=1.0\linewidth]{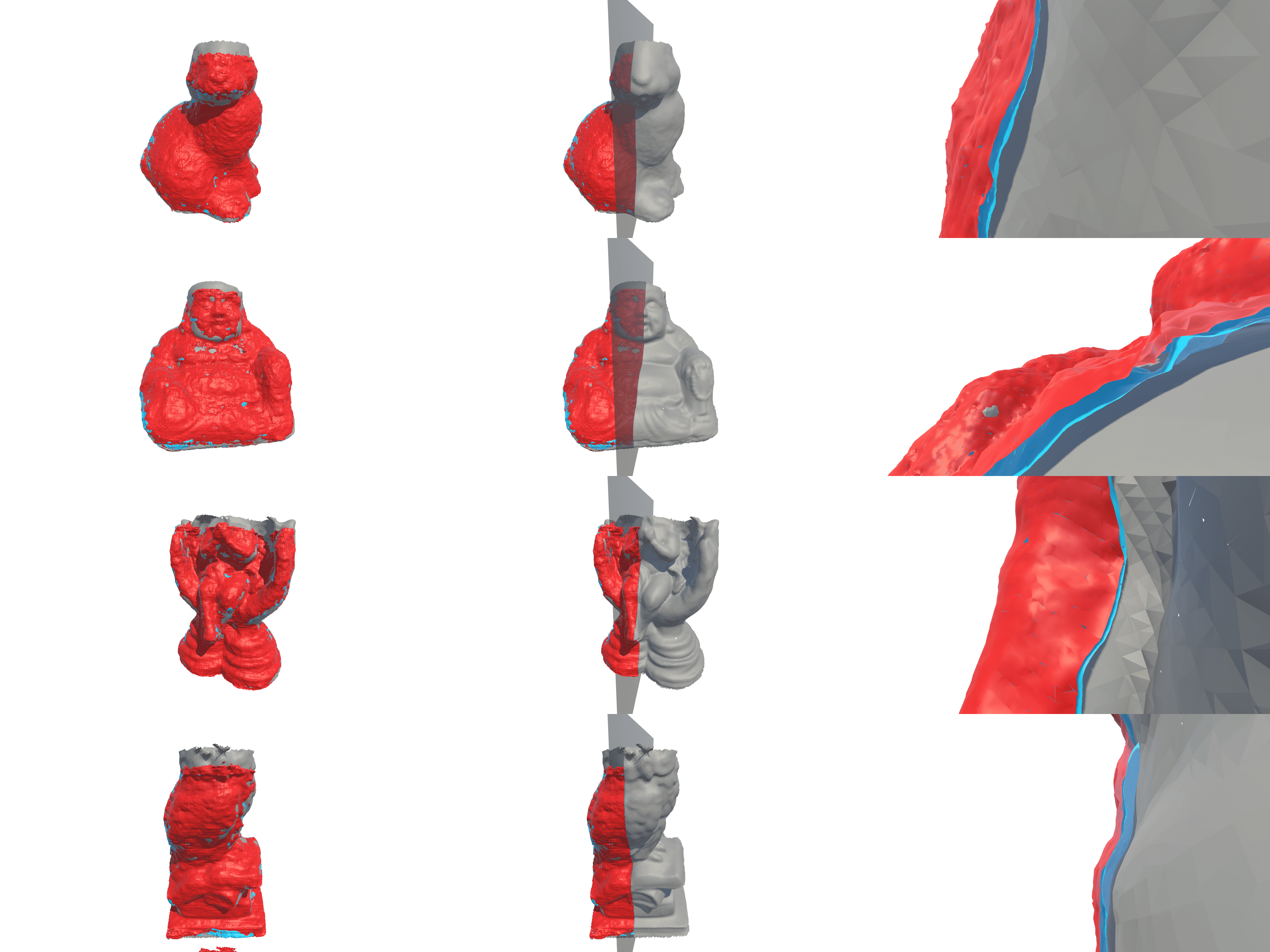} \\ 
    \caption{MVSNeRF \cite{chen2021mvsnerf} (in red) and MVSTransfer (in blue) qualitative evaluation on DTU \cite{aanaes2016large} using $3$ source images. \textbf{Notice that our meshes (in blue) are closer to the ground truth meshes (in gray) than MVSNeRF \cite{chen2021mvsnerf} (in red)}}
    \label{tab:mvsnerf_comparison}
\end{figure*}

As seen above, MVSTransfer (mesh in blue) is closer globally to the ground truth mesh compared to MVSNeRF \cite{chen2021mvsnerf} (mesh in red). This is because of our losses, particularly the weight rendering loss $\mathcal{L}_w(\bold{r})$ which ensures accurate occupancy field estimation in a volumetric rendering framework by reaching a sharp $1$-peak at the location of ray-surface intersection.

\section{Additional ablation studies}
Based on our GeoNeRF experiment \cite{johari2022geonerf}, We ablate the parameters associated with loss $\mathcal{L}_w(\bold{r})$, $a_{max}$ and $a_{min}$ which represent respectively the initial and final widths of the guiding gaussians which 1-peaks at the ray-surface intersection obtained through secant method based root finding. We found the initial value of $a_{max} = 1$ suitable for our training. We ablate the final width $a_{min}$ in Tab.\ref{tab:variance1}. We also ablate the decay parameter $\beta$ in Tab.\ref{tab:variance2}. The latter controls the speed of progression of $a$ in eqn. 14 from $a_{max}$ to $a_{min}$.
\vspace{-10pt}
\begin{table}[h!]
\centering
\begin{minipage}{0.45\linewidth}
\centering
\scalebox{0.7}{
\begin{tabular}{ c|ccc}
 \hline
 $a_{min}$ & 0.002 &0.004 &0.008\\
 \hline
 Chamf. $\downarrow$ & \underline{1.16} & \textbf{1.15} & 1.20\\
 \hline
\end{tabular}
}
\vspace{5pt}
\caption{\footnotesize Ablation of $a_{min}$} \label{tab:variance1}
\end{minipage}
$\quad$
\begin{minipage}{0.45\linewidth}
\centering
\scalebox{0.7}{
\begin{tabular}{ c|ccc}
 \hline
 $\beta$ & 0.0005 &0.001 &0.002\\
 \hline
 Chamf. $\downarrow$ & 1.23 & \textbf{1.15} & \underline{1.17}\\
 \hline
\end{tabular}
}
\vspace{5pt}
\caption{\footnotesize Ablation of $\beta$} \label{tab:variance2}
\end{minipage}
\vspace{-20pt}
\end{table}

These studies demonstrate that our chosen $a_{min}$ and $\beta$ are appropriately chosen to attain accurate generalizable occupancy fields.

\section{Additional qualitative comparison on 3D reconstruction}
Based on our GeoNeRF experiment \cite{johari2022geonerf}, we have included some additional video visualizations of our surface reconstructions in the included supplementary material. There are $4$ on DTU \cite{aanaes2016large}, namely DTU\_Scan$24$.mp4, DTU\_Scan$55$.mp4, DTU\_Scan$118$.mp4 and DTU\_Scan$122$.mp4 and and $2$ on BlendedMVS \cite{yao2020blendedmvs}, namely BMVS\_Scan$3$.mp4, and BMVS\_Scan$12$.mp4.

\section{Inference time}
We use an occupancy representation as it is akin to the sampling and view dependent NeRF opacity under opaque assumption (\cf Unisurf \cite{oechsle2021unisurf}), which facilitates our fast adaptation through transfer learning. Volumetric rendering based inference speed remains similar irrespective of the sdf/occupancy representation. We provide here inference speeds for us and main baselines VolRecon \cite{ren2023volrecon} and ReTR \cite{liang2024retr}. Depth map inference times is about $30$s as reported in their respective supplementary sections. We reproduced this on a RTX A$6000$ and we got $32.4$s for us, $31.8$s for VolRecon \cite{ren2023volrecon} and $37.2$s for ReTR \cite{liang2024retr}.

\section{Additional qualitative comparison on Novel View Synthesis}
\begin{figure*}[t!]
    \centering
    \includegraphics[width=1.025\linewidth]{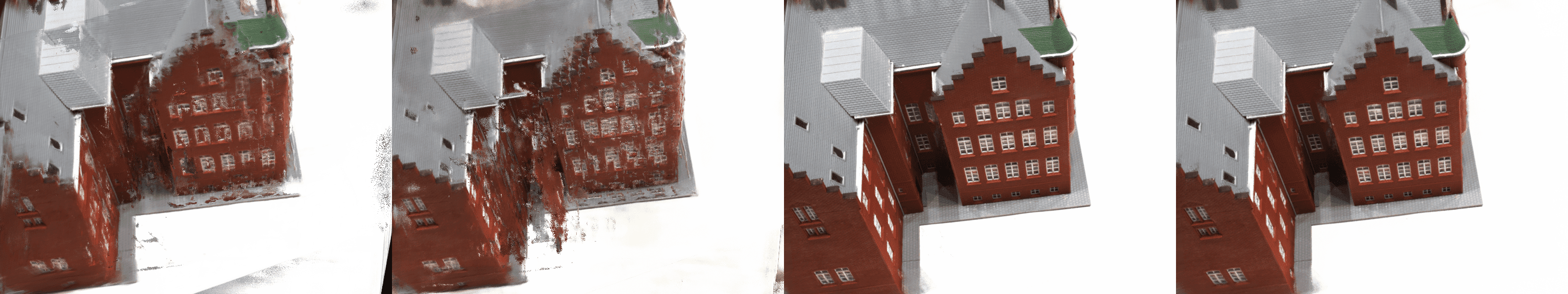} \\ 
    \includegraphics[width=1.025\linewidth]{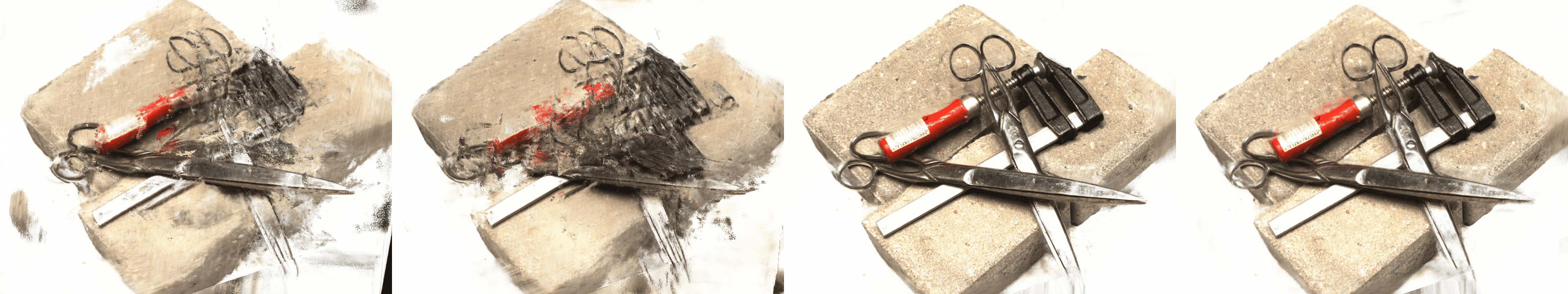} \\ 
    \includegraphics[width=1.025\linewidth]{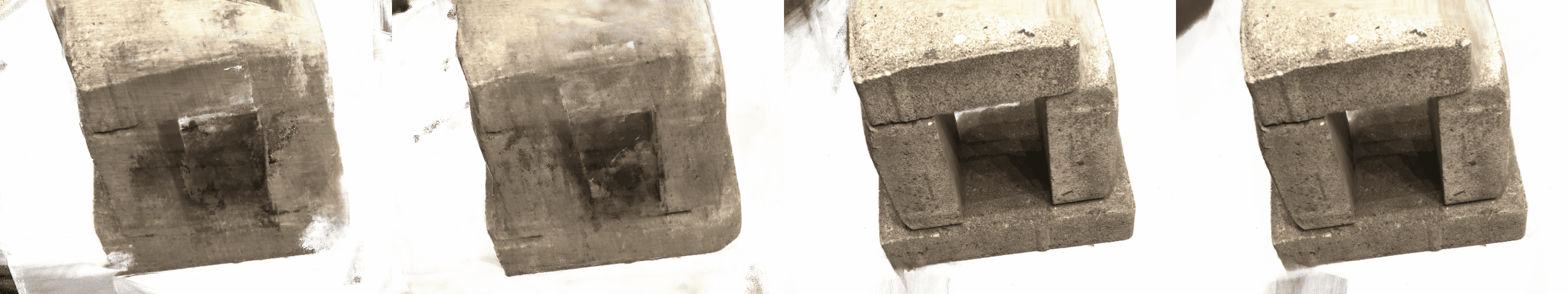} \\ 
    \includegraphics[width=1.025\linewidth]{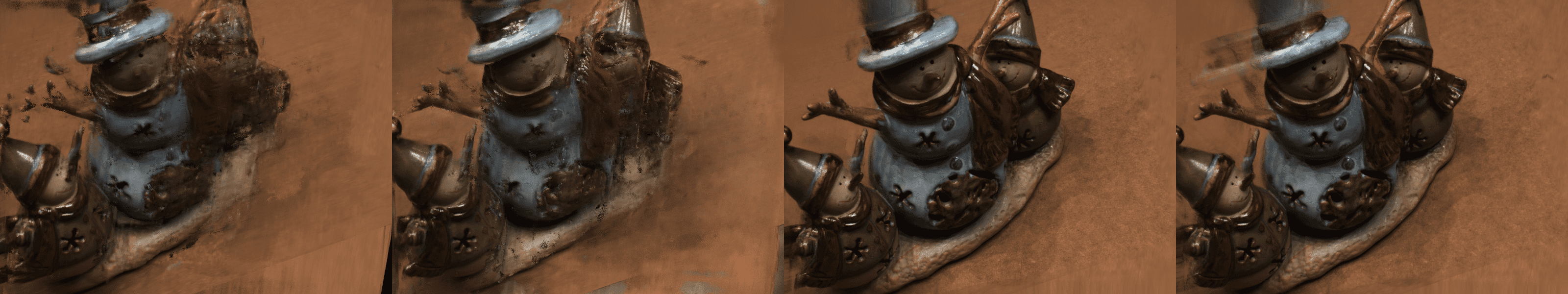} \\ 
    \includegraphics[width=1.025\linewidth]{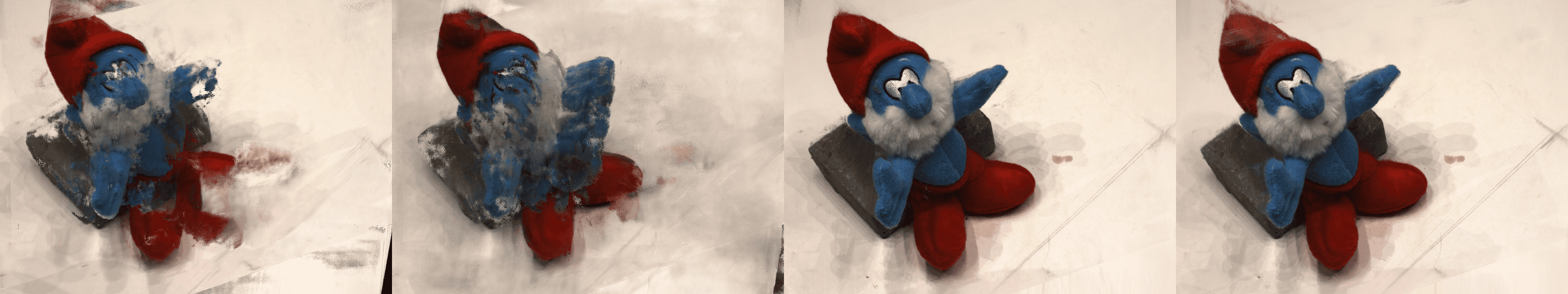} \\ 
    \includegraphics[width=1.025\linewidth]{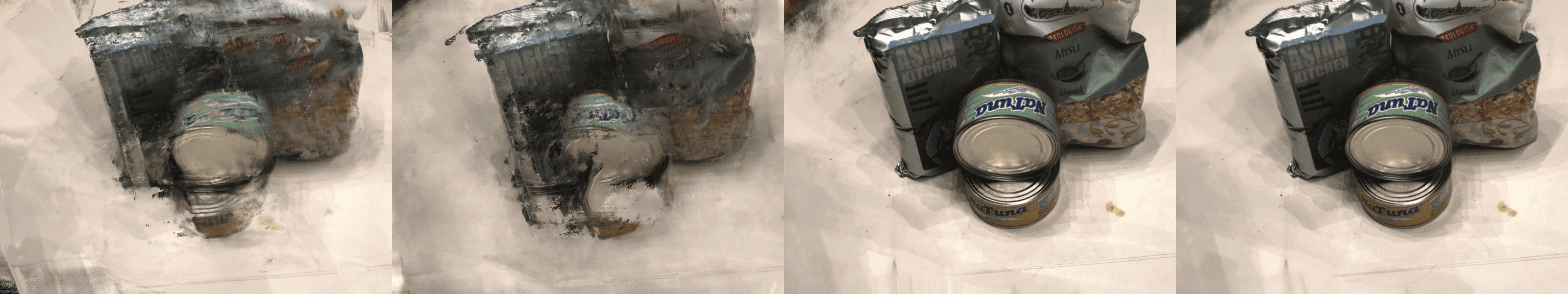} \\ 
    \includegraphics[width=0.825\linewidth]{figs/nvs/Footnote.png} \\ 
    \caption{Novel-View synthesis qualitative evaluation on DTU \cite{aanaes2016large} using $3$ source images.}
    \label{tab:nvs_images1}
\end{figure*}

\begin{figure*}[t!]
    \centering
    \includegraphics[width=1.025\linewidth]{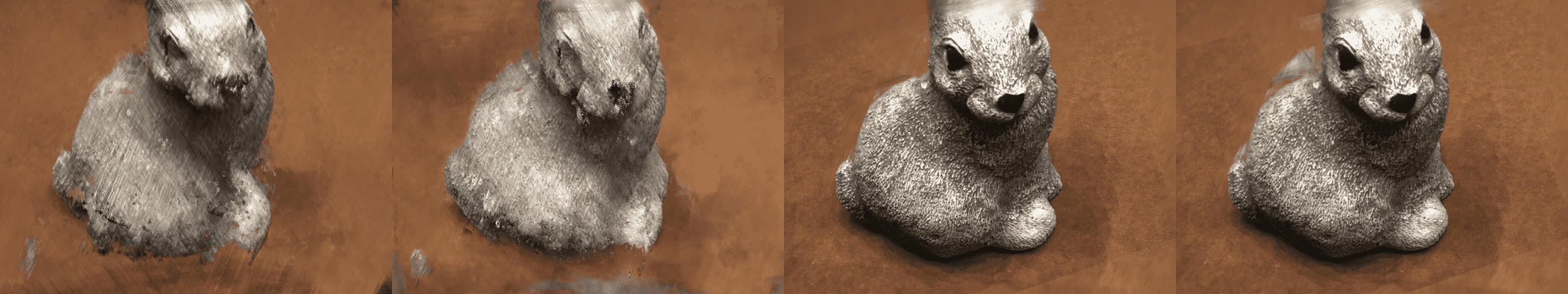} \\ 
    \includegraphics[width=1.025\linewidth]{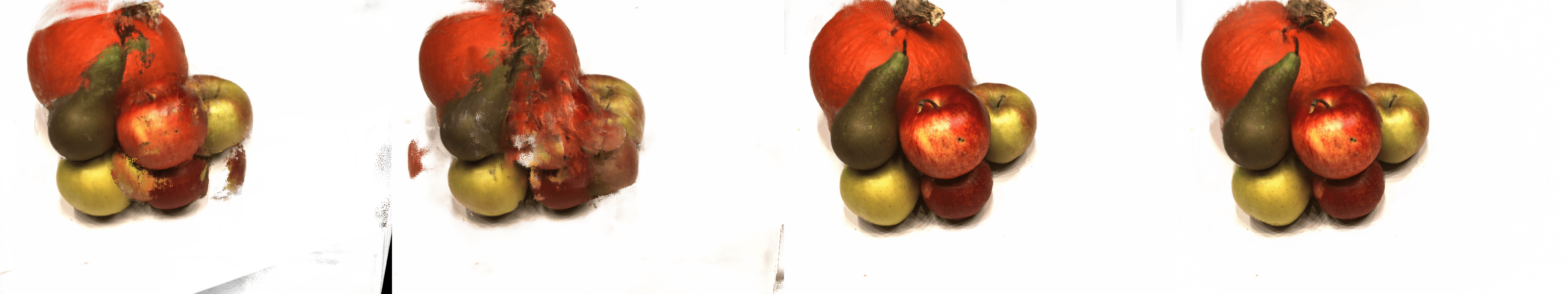} \\ 
    \includegraphics[width=1.025\linewidth]{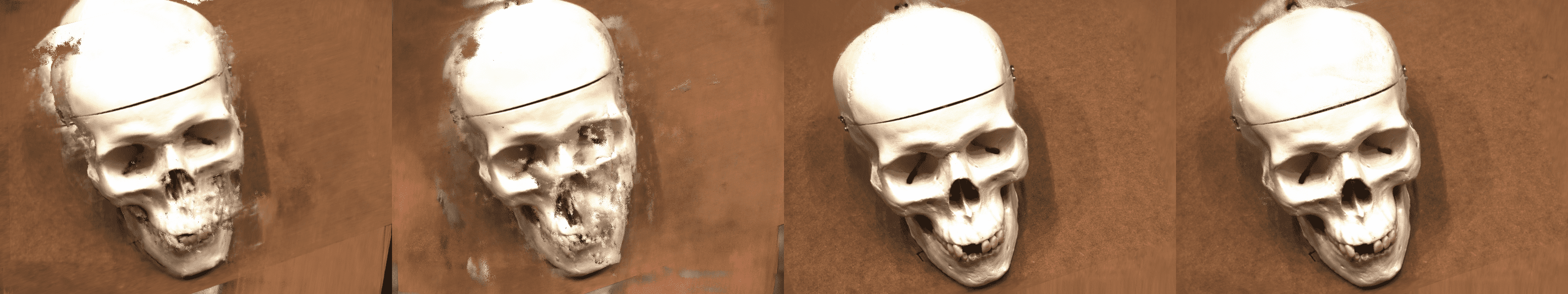} \\ 
    \includegraphics[width=1.025\linewidth]{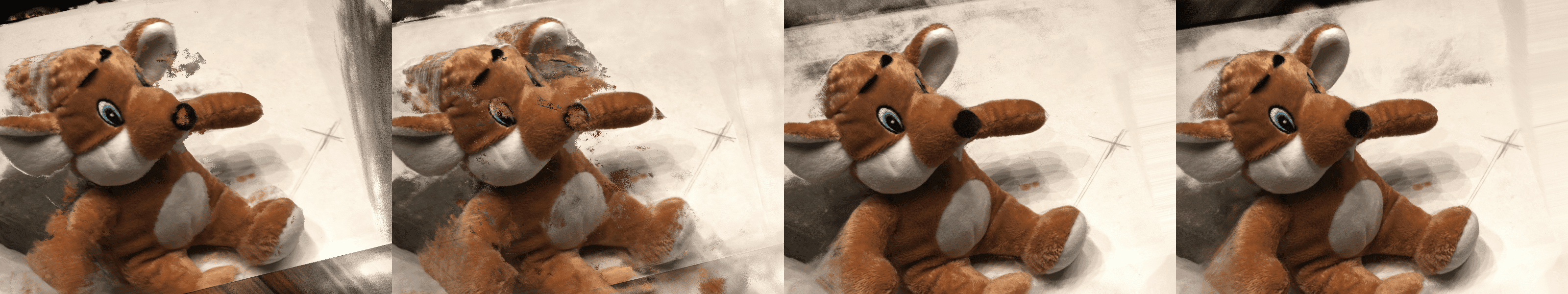} \\ 
    \includegraphics[width=1.025\linewidth]{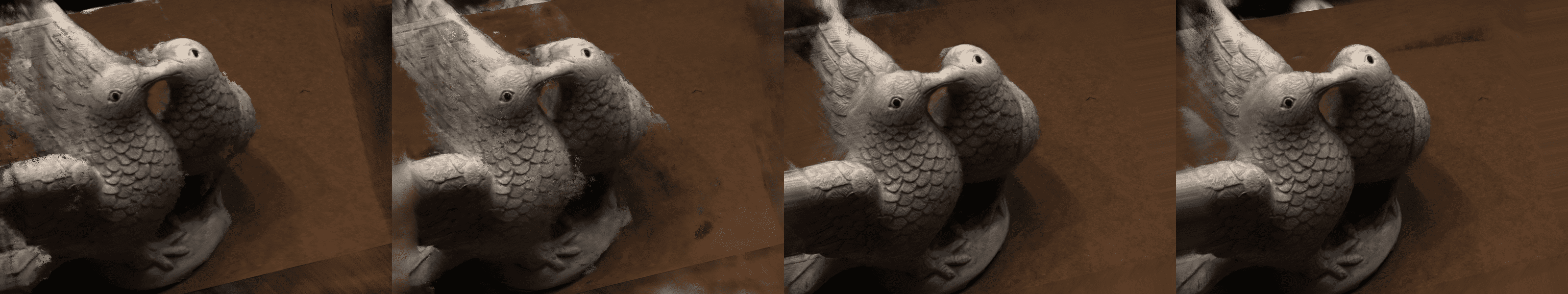} \\ 
    \includegraphics[width=1.025\linewidth]{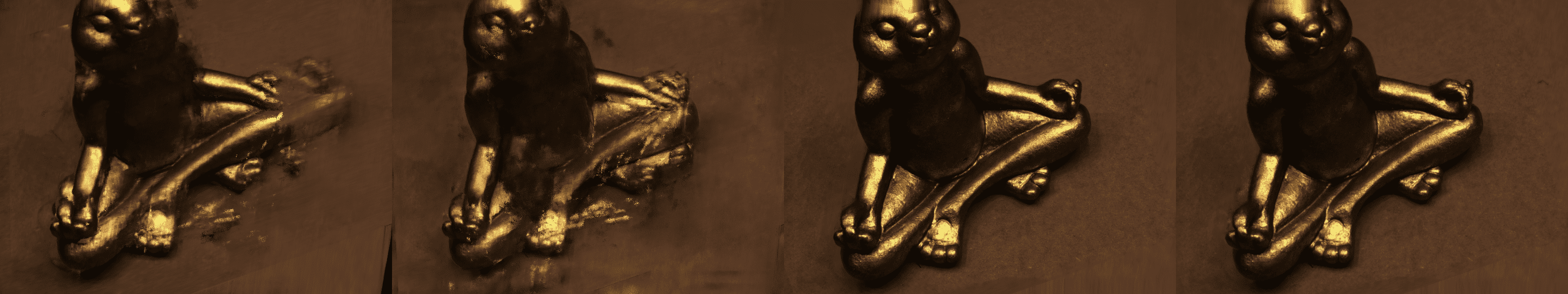} \\ 
    \includegraphics[width=0.825\linewidth]{figs/nvs/Footnote.png} \\ 
    \caption{Novel-View synthesis qualitative evaluation on DTU \cite{aanaes2016large} using $3$ source images.}
    \label{tab:nvs_images2}
\end{figure*}

In this section, based on our GeoNeRF experiment \cite{johari2022geonerf}, we present additional qualitative comparisons with VolRecon \cite{ren2023volrecon} and ReTR \cite{liang2024retr} in Figure \ref{tab:nvs_images1} and \ref{tab:nvs_images2} to demonstrate the superior performance of our method on the DTU \cite{aanaes2016large} dataset. Our final adapted model preserves the novel-view capabilities of its initial backbone and provides good novel-view extrapolation results compared to the generalizable reconstruction networks (\eg VolRecon \cite{ren2023volrecon} and ReTR \cite{liang2024retr}). We notice that qualitatively, in the sparse $3$ input views setting, we are sharper than the competing methods, with lesser artifacts. This demonstrates the robustness of our method on the task of novel-view synthesis, apart from also displaying \textbf{state-of-the-art} results on surface reconstruction. 

\clearpage

\clearpage

\section{Additional Experimental Details}
\label{sec:exp_details}
In this work, we evaluated two sets of data: DTU \cite{aanaes2016large} and BlendedMVS \cite{yao2020blendedmvs}. For DTU \cite{aanaes2016large}, we follow distinct protocols based on the task's nature, distinguishing between novel view synthesis and surface reconstruction.

\paragraph{\textbf{Metrics}} For the novel view synthesis task involve evaluating PSNR scores, assuming a maximum pixel value of 1 and using the formula $-10 \log_{10}$(MSE). Additionally, we employ the scikit-image's API to calculate the Structural Similarity Index (SSIM) score and the pip package lpips, utilizing a learned VGG model for computing the Learned Perceptual Image Patch Similarity (LPIPS) score. In the context of the surface reconstruction task, we gauge Chamfer Distances by comparing predicted meshes with the ground truth point clouds of DTU scans. The evaluation process follows the methodology employed by SparseNeuS, VolRecon, ReTR \cite{long2022sparseneus, ren2023volrecon, liang2024retr}, employing an evaluation script that refines generated meshes using provided object masks. Subsequently, the script evaluates the chamfer distance between sampled points on the generated meshes and the ground truth point cloud, producing distances in both directions before providing an overall average, typically reported in evaluations. Additionally, two sets of 3 different views are used for each scan, and we average the results between the two resulting meshes from each set of images and report it in the comparison as done in previous methods \cite{long2022sparseneus, ren2023volrecon, liang2024retr}.

\paragraph{\textbf{DTU Dataset}} The DTU dataset \cite{aanaes2016large} is an extensive multi-view dataset comprising $124$ scans featuring various objects. Each scene is composed of $49$–$64$ views with a resolution of $1600 \times 1200$. We adhere to the procedure outlined in \cite{long2022sparseneus, ren2023volrecon, liang2024retr}, training on the same scenes as employed in these methods and then test on the $15$ designated test scenes for both the reconstruction and novel view synthesis tasks. The test scan IDs for both novel view synthesis and surface reconstruction are : $24$, $37$, $40$, $55$, $63$, $192$, $65$, $69$, $83$, $97$, $105$, $106$, $110$, $114$, $118$ and $122$. For surface reconstruction, for each scan, there are two sets of 3 views with the following IDs used as the input views: set-$0$: $23$, $24$ and $33$, then set-$1$: $42$, $43$ and $44$ all scans. We use the training views in half resolution, \ie $800 \times 600$. As for novel-view synthesis, we test for camera views not used during training for a fair evaluation, with the following IDs used as the target and source views for all the scenes - $37$ : $39$, $36$ and $20$; $38$ : $39$, $37$ and $40$; $39$ : $40$, $37$ and $36$.  

\paragraph{\textbf{BlendedMVS Dataset}} BlendedMVS \cite{yao2020blendedmvs} is a large-scale dataset for generalized multi-view stereo that consists of a variety of $113$ scenes including architectures, sculptures and small objects with complex backgrounds. For surface-reconstruction, we use $7$ challenging scenes, in accordance with SparseNeuS \cite{long2022sparseneus} where each scene has $31$–$143$ images captured at $768 \times 576$. The chosen IDs for the selected scenes are : Scan$2$ : $67$, $29$, $59$; Scan$3$ : $1$, $0$, $2$; Scan$12$ : $2$, $8$, $50$; Scan$13$: $28$, $4$, $11$; Scan$14$: $9$, $109$, $50$; Scan$22$: $4$, $3$, $5$; Scan$24$: $23$, $39$, $5$. We use the testing views in their original resolution.

\paragraph{\textbf{Masked evaluation for novel view synthesis}} In accordance with the findings presented by \cite{niemeyer2022regnerf} addressing the bias in background evaluation, we adopt their prescribed approach of masked evaluation for DTU \cite{aanaes2016large}. This involves employing object masks and computing PSNR exclusively within the defined mask. In the case of SSIM and LPIPS, we utilize the masks to superimpose the predicted object-of-interest onto a black background before metric calculations.
\clearpage  

%
%
\bibliographystyle{splncs04}
\bibliography{main}

\end{document}